\documentclass{article}
\usepackage{microtype}
\usepackage{spconf,amsmath,graphicx}
\usepackage{xcolor}
\usepackage{balance}
\usepackage{hyperref}

\usepackage[ruled,vlined]{algorithm2e}
\DeclareMathOperator*{\argmax}{argmax} % thin space, limits underneath in displays

\usepackage{multirow}

\title{Headless Horseman: Adversarial Attacks on Transfer Learning Models}
\name{Ahmed Abdelkader$^1$, Michael J. Curry$^1$,  Liam Fowl$^2$, Tom Goldstein$^1$}
\secondlinename{Avi Schwarzschild$^2$, Manli Shu$^1$, Christoph Studer$^3$, Chen Zhu$^1$}
\address{$^1$Dept. of Computer Science, $^2$ Dept. of Mathematics, University of Maryland, $^3$ Cornell Tech}

\begin{document}
%\ninept
%
\maketitle
\begin{abstract}
Transfer learning facilitates the training of task-specific classifiers using pre-trained models as feature extractors.  We present a family of transferable adversarial attacks against such classifiers, generated without access to the classification head; we call these \emph{headless attacks}.  We first demonstrate successful transfer attacks against a victim network using \textit{only} its feature extractor. This motivates the introduction of a label-blind adversarial attack. This transfer attack method does not require any information about the class-label space of the victim. Our attack lowers the accuracy of a ResNet18 trained on CIFAR10 by over 40\%. %77.9 to 33.8

% A simple method of generating black-box adversarial examples is with surrogate networks. We show that a cleaver choice of surrogate leads to a stronger attack. In particular, we use meta-learned models and find that they lead to stronger attacks than standard trained models of the same architectures. \todo{And even than more complicated architectures whose parameters were not meta-learned.}
\end{abstract}
\section{Introduction}
\label{sec:intro}
Neural networks are powerful tools for solving computer vision problems, but training them from scratch requires huge amounts of data and compute time \cite{he2016deep, szegedy2016rethinking}.  One of the most popular frameworks for reducing data requirements is {\em transfer learning} \cite{yosinski2014transferable} in which a pre-trained network (usually trained on a large labeled dataset like ImageNet) is used to extract low-dimensional features from images. A new linear classifier head is then trained to classify images using a small task-specific dataset and the corresponding number of outputs.  Transfer learning is widely used in practice thanks to the availability of standard pre-trained models within common deep learning frameworks like PyTorch and TensorFlow.

In this paper, we study the security vulnerabilities introduced by simple transfer learning strategies.  Deep neural networks are known to be vulnerable to \textit{adversarial attacks}: inputs that have been maliciously crafted to fool a victim network.   These attacks generally follow two paradigms: (i) white-box attacks, in which the attacker has complete knowledge of the victim network (both the network architecture and weights) and (ii) black-box attacks in which the attacker does not know the victim network but can query its output label on chosen inputs \cite{papernot2016practical,goodfellow_explaining_2014}.

We study a new family of ``headless'' attacks, in which the attacker only has knowledge of the feature extractor being used, but no knowledge of the classifier head and no access to its output.  This is a realistic threat model in situations where the victim network uses a standard pre-trained model, such as those that ship with the PyTorch distribution,\footnote{\url{https://pytorch.org/docs/stable/torchvision/models.html}} or those available for download from other popular GitHub repositories.  

We find that using a known feature extractor exposes a victim to powerful attacks that can be executed without knowledge of the classifier head at all.  In particular, an attacker can successfully mount an adversarial attack with no knowledge of the task-specific dataset used by the victim, the class labels used by the task-specific head, or even the number of classes in the victim's training set.

\section{Related work}
\label{sec:relwork}
We review related work on transfer learning and adversarial attacks, highlighting the most closely related attacks to the threat model we consider.

\subsection{Fine-tuning and transfer learning}

With the introduction of very large datasets like ImageNet, and the use of networks with dramatically increased depth, neural networks have become increasingly successful for many applications \cite{he2016deep,deng2009imagenet}. However, the size of datasets and depth of networks means that training models to state-of-the-art quality may be infeasible without immense computational resources. Since neural networks can learn feature representations of the input data that are generically useful \cite{pmlr-v27-bengio12a}, it is possible to use them as feature extractors for other tasks \cite{mao1995artificial}. 
Formally, we start with a pre-trained feature extractor $f$ with parameters $\phi$ and optimize the head's parameters, $\omega$, as 
\[ \min_\omega \mathcal{L} (f(\mathbf{x}; \phi), y ; \omega), \] 
where $\mathcal{L}$ is the standard cross-entropy loss. Note that only the parameters of the head ($\omega$, often just a linear layer) are trained in this setup. 
%Large, well-resourced organizations often train very deep models on very large datasets and provide the trained models for free -- then end-users can fine-tune their own models by training only the last layer for their specific problems.

\subsection{Adversarial attacks}

Adversarial attacks on deep neural networks are typically considered in either the \textit{white-box} or \textit{black-box} settings.

In the white-box setting, the attacker is assumed to have access to the model to be attacked, including arbitrary gradients \cite{goodfellow_explaining_2014}. An early, simple white-box attack, called Fast Gradient Sign Method (FGSM) makes a bounded $\ell_\infty$ perturbation according to the sign of the gradient of the network's loss function with respect to the input \cite{goodfellow_explaining_2014}. A more powerful attack, known as projected gradient descent (PGD), iteratively crafts a perturbation by maximizing the network loss in a bounded $\epsilon$-ball around the input \cite{kurakin2016adversarial, aleks2017deep}. Other attacks include DeepFool \cite{DeepFool}, which assumes a linear network to calculate the perturbation, or the Carlini-Wagner attack \cite{Carlini_2017}, which uses projected gradient descent on an unconstrained loss function.

In the black-box setting, the attacker can only query the victim and may not know its parameters or even its architecture \cite{papernot2016practical}. One variant of a black-box attack trains a surrogate network on the same data on which the victim network was trained, and crafts perturbations on the surrogate network in the hope that the perturbation will transfer to cause misclassification bt the victim network \cite{PMG_transferability}. Other variants try to estimate the gradient of the network using softmax outputs of the network to craft an attack using standard white-box methods \cite{ilyas2018prior, bhagoji2017exploring}. On the theory side, some work has been done explaining the transferability of black-box attacks via the implicit regularization of model complexity \cite{demontis2019adversarial}.  

In this paper, we consider a threat model which is different from both of these models, while combining their properties. Our threat model assumes that the victim's feature extractor is known to the attacker, but the last layer, including the class-label space, is unknown to the attacker. Unlike both the white-box and black-box models, we do \textit{not} assume the attacker knows which classes or how many classes the victim model is designed to classify. This is motivated by the widespread use of transfer learning. For many tasks where labeled data is sparse, users will take an expressive, pre-trained feature extractor, e.g., the convolutional layers of ResNet, and fine-tune a classification head on a new task to quickly deploy a network on this task \cite{he2016deep, yosinski2014transferable}.

\subsection{Other transfer-based attacks}
Other works have considered adversarial attacks where the classification head is not available and the feature extractor is pre-trained. An attack model with stronger assumptions was considered in \cite{wang2018transfer}, where the attacker is assumed to have partial knowledge about the labels of some input images and exploits that knowledge to simply move the input features close to a target image of a different label. The grey-box model considered in \cite{prabhugrey} is closely related to our work, but was only considered in a very limited setting with two class labels and no successful attack was demonstrated.

\section{Headless Attacks}

We first test the importance of the classification head in crafting adversaries with a head-agnostic attack. These experiments motivate our \textit{label-blind} attack wherein the attacker has no knowledge of the victim network's architecture or parameters. In this setting, the attacker also has no access to a labeled training dataset (and therefore does not know the number of classes and the distribution of labels for the victim's dataset). 

\subsection{Centroid-based attack}
To test the importance of the classification head, we construct a transfer attack while denying access to the classification head. We attack only the feature extractor of a surrogate network using feature perturbations. We make use of the distances to the class centroids as synthetic prediction logits, replacing those that would have been output by the classification head. Specifically, we consider the following optimization problem for the perturbation $\delta$:
\begin{equation}
    \label{eqn:centroidattack}
    \argmax_{||\delta||_\infty \leq \epsilon} 
    \mathcal{L}(\mathbf{d}(x + \delta),\text{OneHot}(y)), \\
\end{equation}
where \[
   (\mathbf{d}(x))_{y} = \lVert f(\mathbf{x}) - \mu_{y} \rVert \: \text{ and } \:\:
    \mu_{y} = \frac{1}{N}\sum_{(\mathbf{x},y') \text{ s.t. } y' = y} f(\mathbf{x}),
\]
with $\mathcal{L}$ denoting the standard cross-entropy loss and $f$ denoting the feature extractor.

We find that this ``headless'' centroid-based transfer attack is competitive with PGD attacks that have access to the surrogate prediction head (see Table \ref{tab:centroid}). This serves as evidence that having access to just the features, and not the logits of a network, is sufficient for constructing adversarial attacks.

\subsection{Label-blind attack}

We now consider the setting where the attacker only knows the feature extractor that is used, and knows nothing about the victim's dataset, \emph{including the number of classes and class-labels}. We call this setting the \textit{label-blind} setting. Given an image of interest for the victim, the goal is to generate a perturbation that will cause misclassification by the victim.

We compute the perturbation using the output of a pre-trained ImageNet classifier as synthetic targets for an ordinary PGD attack. We do not assume the input was used to train the ImageNet classifier, or that the image lies within an ImageNet class. Rather, we rely on the ImageNet classifier being able to extract discriminative features from the input. We then find a perturbation that causes these features to shift toward a centroid further away than the nearest class centroid. We show that this perturbation transfers to other classifiers that use this same feature extractor and thus cause these networks to make errors. Our label-blind attack is detailed in Algorithm \ref{headless_alg}.

In general, we select a certain logit for each sample according to the ranking ($i$) of that logit among all logits.
Then we optimize the perturbation to minimize the cross entropy loss on logit $i$, such that the perturbed image in the feature space of this extractor lies closer to the manifold of a certain class different from its ground truth. 

\SetArgSty{textnormal}
\begin{algorithm}[h!]
$\text{\bf{Require:}}$ Surrogate network $S$, image $\mathbf{x}$, perturbation $\delta$, logit ranking index $i$, number of steps per PGD attack $K$, and maximum attack radius $\epsilon$.\\
%\SetAlgoLined

  Initialize $\delta \in \mathcal{B}_\epsilon (\mathbf{x})$ randomly\; 
  \For{$step$ = 1,...,$K$}{
   \textbf{Calculate} cross-entropy loss between the output of the surrogate network and the one-hot vector corresponding to the $i^{th}$ highest logit in the surrogate network's output on the clean data: 
    \[
    \mathcal{L}\big(S(\mathbf{x} + \delta), \text{OneHot}(i) \big) \]
   
   \textbf{Compute} $g = \text{sign} \left( \nabla_\mathbf{\delta} \mathcal{L}\big(S(\mathbf{x} + \delta), \text{OneHot}(i) \big) \right)$. \\
      \textbf{Update} $\delta = \delta - \gamma g$. \\
      If $\|\delta\|_\infty>\epsilon$, then project $\delta$ onto the ball $\mathcal{B}_\epsilon (\mathbf{x})$.
     }
     \textbf{return} perturbed image $\mathbf{x} + \delta$

\label{headless_alg}
\caption{Headless Horseman Algorithm}
\end{algorithm}

\section{Experiments}
\label{sec:experiments}
Our experimental results are evaluated on the CIFAR-10 dataset \cite{krizhevsky2014cifar}.
The networks trained on CIFAR-10 follow the default settings (including the mean and standard deviation for normalizing the images in all our experiments) from a published repository.\footnote{https://github.com/kuangliu/pytorch-cifar.git} 
The pre-trained ImageNet feature extractors are provided by PyTorch. Our code repository is publicly available online.\footnote{https://github.com/zhuchen03/headless-attack.git}

\subsection{Centroid-based attacks}

For our centroid-based attacks, we experiment with a variety of pre-trained networks (trained with different random seeds) for the victim \cite{Sandler_2018, hu2018squeeze}. We compute the distance between the target image's feature representation and the centroids of all the other classes in feature space. Using these distances as synthetic logits we minimize the cross-entropy loss for the ground-truth class. It is critical to observe that the synthetic logits are large for classes whose feature-space centroid is far away, and therefore minimizing the cross-entropy loss perturbs the feature representation in an adversarial direction, which maximizes the distance to the ground truth class and minimizes the distance to all other classes. We also project the perturbed example in input space onto an $\ell_\infty$ $\epsilon$-ball. We consider four choices of $\epsilon$.
 
The results of experiments with centroid-based attacks are summarized in Table \ref{tab:centroid}. For most of the $\epsilon$ values, the centroid-based attack outperforms the transferred PGD attack by a small amount; for $\epsilon=4/255$ we see better performance from the transferred PGD attack. 
In all cases, however, performance is reasonably close, suggesting that the centroid-based attack (which requires no knowledge of the classifier head) is a viable alternative to transferred PGD (which requires such knowledge). We obtain comparable results transferring from ResNet18 to other architectures as shown in Table \ref{tab:othermodels}.

\begin{table}[tp]
\centering
    \begin{tabular}{|c|c|c|}
    \hline
        $\epsilon$  & Centroid & PGD \\
        \hline
        \hline
         1/255 & \bf{85.13 (.05)} & 85.75 (.09) \\
         2/255 & \bf{66.87 (.09)} & 67.41 (.12) \\
         4/255 & 37.64 (.20) & \bf{35.24 (.38)} \\
         8/255 & \bf{10.57 (.20)} & 11.01 (.29) \\
         \hline
    \end{tabular}
    \caption{Mean and standard deviation of accuracy of classifiers under Centroid and PGD attacks. All results are based on 20-step attacks with a step size of 0.05 in the normalized image space. Percentages are averaged over 5 runs. The bold numbers indicate better results. Both the surrogate and victim networks are ResNet18 models.}
\label{tab:centroid}
\end{table}

\begin{table}[tp]
\centering
\begin{tabular}{|c|c|c|c|c|}
\hline
\multirow{2}{*}{$\epsilon$} & \multicolumn{2}{c|}{MobileNetv2~\cite{Sandler_2018}} & \multicolumn{2}{c|}{SENet18~\cite{hu2018squeeze}} \\ \cline{2-5} 
                                         & Centroid         & PGD           & Centroid       & PGD         \\ \hline
                                         \hline
1/255                                    & \textbf{81.08}           & 81.58        & \textbf{85.27}         & 86.03      \\ \hline
2/255                                    & \textbf{70.81}           & 72.56        & \textbf{73.11}         & 74.16      \\ \hline
4/255                                    & %\textbf{53.02}           & 55.18        & 52.60          & \textbf{5.070}       \\ \hline
\textbf{53.02}           & 55.18        & 52.60          & \textbf{50.70}       \\ \hline
8/255                                    & \textbf{26.35}           & 29.91        & 23.74         & \textbf{22.39}  \\
\hline
\end{tabular}
\caption{Performance of centroid and PGD attacks transferred from ResNet18 to other architectures, for single trials.}
\label{tab:othermodels}
\end{table}

% \begin{table*}[t]
% \centering
% \begin{tabular}{|c|c|c|c|c|c|c|c|c|c|c|}
% \hline
% \multirow{2}{*}{\epsilon} & \multicolumn{2}{c|}{MobileNet}  & \multicolumn{2}{c|}{VGG16}        & \multicolumn{2}{c|}{ResNeXt29}  & \multicolumn{2}{c|}{ResNet50}   & \multicolumn{2}{c|}{DPN26}      \\ \cline{2-11} 
%                                          & Centroid       & PGD            & Centroid       & PGD            & Centroid       & PGD            & Centroid       & PGD            & Centroid       & PGD            \\ \hline
% 1/255                                    & \textbf{81.87} & 82.36          & \textbf{89.16} & 89.53          & \textbf{91.71} & 91.90          & \textbf{87.45} & 88.15          & \textbf{87.41} & 88.10          \\ \hline
% 2/255                                    & \textbf{71.45} & 72.92          & \textbf{82.92} & 83.20          & 84.85          & \textbf{84.58} & \textbf{77.69} & 78.88          & \textbf{75.97} & 77.13          \\ \hline
% 4/255                                    & \textbf{52.62} & 52.66          & 71.19          & \textbf{66.66} & 67.53          & \textbf{63.53} & 60.91          & \textbf{58.82} & 56.58          & \textbf{53.31} \\ \hline
% 8/255                                    & 25.74          & \textbf{25.66} & 43.37          & \textbf{38.00} & 33.05          & \textbf{31.00} & 33.46          & \textbf{32.38} & \textbf{25.51} & 26.55          \\ \hline
% \end{tabular}
% \end{table*}

\subsection{Label-blind attacks}
To test our algorithm in the label-blind transfer learning setting, we use a ResNet50 pre-trained on ImageNet. We freeze all the convolutional layers and only train the final fully connected layer. In order to match input dimensions, CIFAR-10 images are up-sampled (we did not use other data augmentation). This network becomes the victim model.  
Such a model achieves 77.90\% accuracy on the up-sampled CIFAR-10 test set.
Then, we apply the label-blind attack on the same pre-trained ResNet18 feature extractor.

\begin{figure}[tb]
\center{\includegraphics[width=0.5\textwidth]{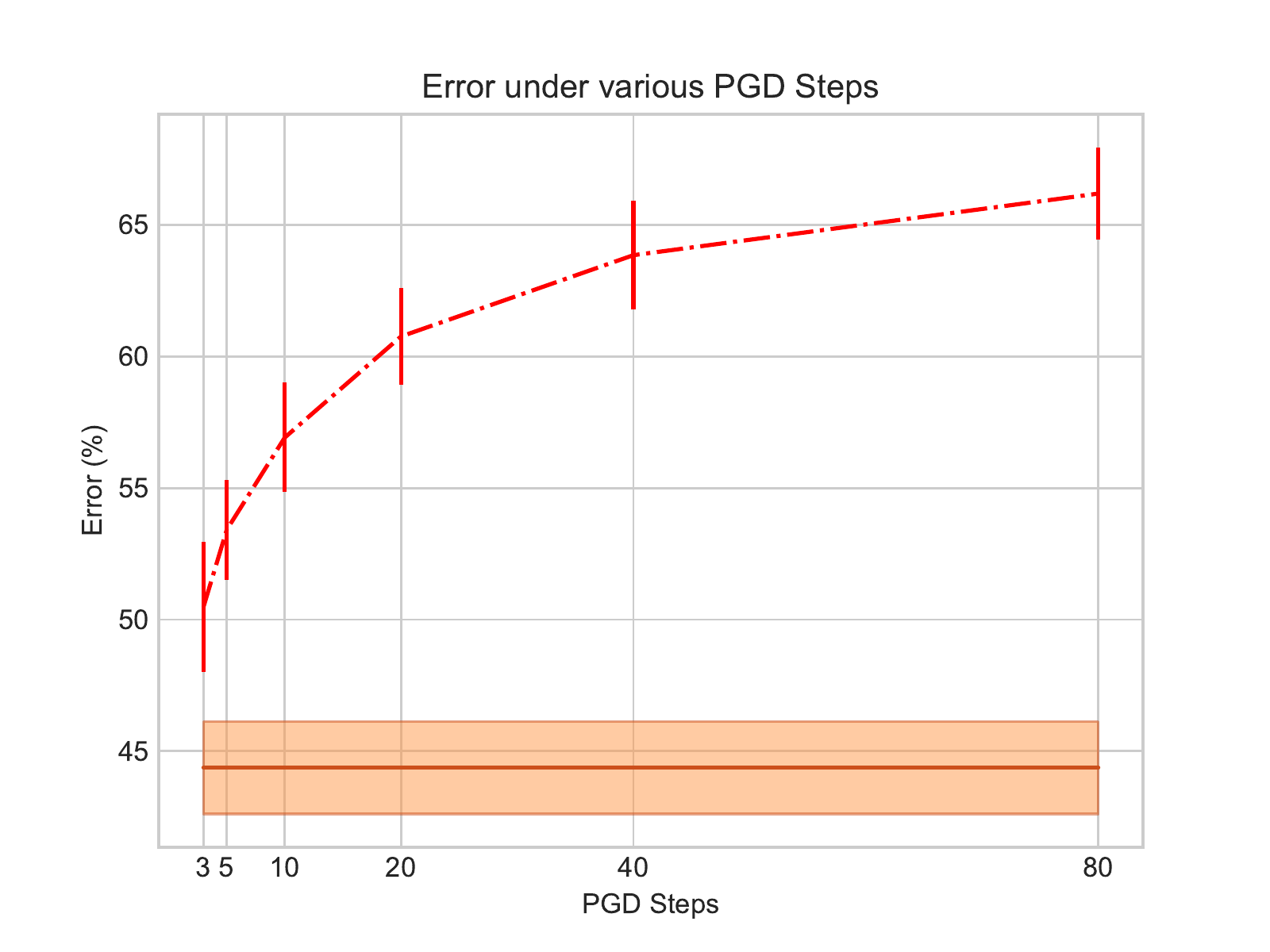}}
\caption{\label{fig:steps} Mean and standard deviation of the error of the transfer-learned model under different numbers of PGD steps. We fix the step size of PGD to 0.05 in the normalized image space, and use $\epsilon=8/255$. The underlying line shows the mean and standard deviation of the error under Gaussian random perturbation with the same magnitude. All results are based on 4 runs with different random initial perturbations.}
\end{figure}

Fig.~\ref{fig:steps} shows the error of the victim model under various steps of attacks. 
Under 80-step label-blind PGD attack, the error of the victim model on CIFAR-10 test set is increased from 22.10\% to 66.19\%. 
In contrast, adding random Gaussian noise of the same magnitude to the images results in an error of 44.38\%. 
For every number of steps we try, the error after our attack is higher than random noise plus its standard deviation, demonstrating the effectiveness of the Headless Horseman attack.

\begin{figure}[tb]
\center{\includegraphics[width=0.5\textwidth]{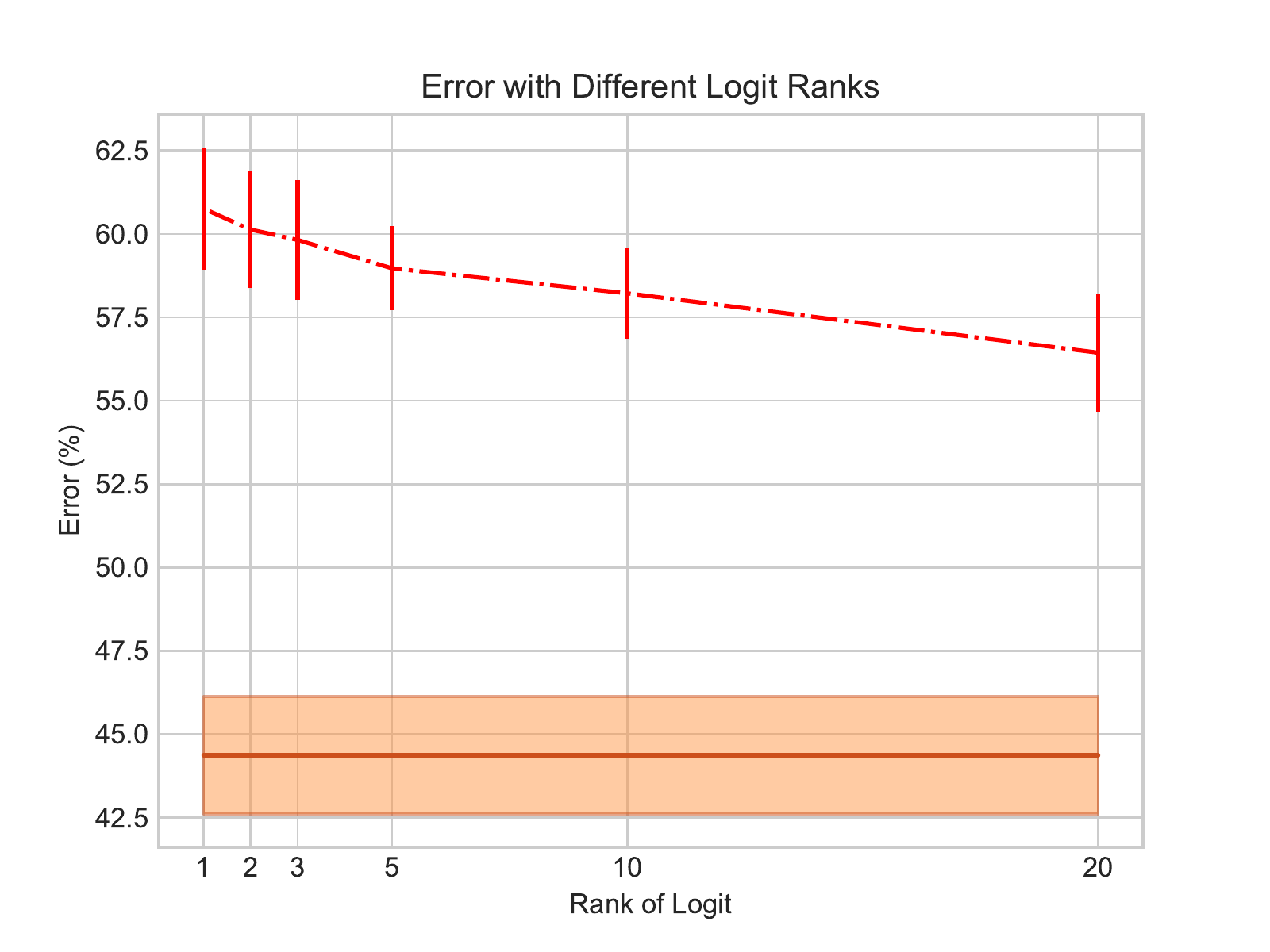}}
\caption{\label{fig:labelrank} Mean and standard deviation of the error of the transfer-learned model when choosing logits with different rankings. All results are with 20-step iterative attacks. Step size and $\epsilon$ are the same as in Fig.~\ref{fig:steps}. The results are based on 4 runs. }
\end{figure}

Fig.~\ref{fig:labelrank} shows the error of the victim model for different choices of logit rankings $i$.
% We perform a search over the logit ranking in the Headless Horseman algorithm in figure~\ref{fig:labelrank}. 
We find that targeted attacks, which target more likely classes, transfer more successfully than when the target classes are chosen as less likely labels. 
This suggests that the features that can be most easily perturbed to cause misclassification on the surrogate network are also the features that are most easily perturbed to cause misclassification on the victim network, even though the surrogate network was trained on a different dataset. 

\section{Discussion}
\label{sec:discussion}

We consider adversarial attacks in the ``headless'' setting where the victim is using a public, pre-trained feature extractor, known to the attacker, but the classification head is not available. This is a realistic setting -- fine-tuning of classifiers on pre-trained networks is a widespread practice. We consider two variations of this setting.

In the first setting, the attacker knows the dataset that the victim is trained on, and aims to perform a transfer attack using a surrogate model. In this setting, we find that performing a ``headless'' centroid-based attack which ignores the classification layer performs competitively with a PGD attack, which requires access to the surrogate's logits. Success in this setting motivates our second attack.

In the second setting, the attacker knows nothing about the dataset of interest to the victim, not even the number of labels, however, they do know that the victim has used a particular pre-trained model. This setting is especially realistic since many networks use common, pre-trained feature extractors for previously unlearned tasks. This setting is largely ignored by standard white-box and black-box adversarial attacks. In this setting, given a test image, the attacker can craft an adversarial perturbation with the feature extractor and whichever linear classification head is provided with the network. This attack transfers to a victim network that uses the same feature extractor.
% A new linear classifier head is then trained to classify images using a small task-specicific dataset and the corresponding number of outputs.

\section{Conclusions}
One reason for the success of transfer learning is the expressive feature extractor's ability to discern discriminative features. However, this characteristic also makes transfer-learned models vulnerable to transferable attacks that require no knowledge of the dataset being used, or of the label space. We hope this work raises awareness of such security vulnerabilities, and expect these results to encourage practitioners to avoid the practice of using publicly-available pre-trained networks for sensitive applications without adequate precautions.

\section{Acknowledgements}
This work was supported by the DARPA GARD, DARPA QED4RML programs, and National Science Foundation DMS division.

% Below is an example of how to insert images. Delete the ``\vspace'' line,
% uncomment the preceding line ``\centerline...'' and replace ``imageX.ps''
% with a suitable PostScript file name.
% -------------------------------------------------------------------------
% \begin{figure}[htb]

% \begin{minipage}[b]{1.0\linewidth}
%   \centering
%   \centerline{\includegraphics[width=8.5cm]{image1}}
% %  \vspace{2.0cm}
%   \centerline{(a) Result 1}\medskip
% \end{minipage}
% %
% \begin{minipage}[b]{.48\linewidth}
%   \centering
%   \centerline{\includegraphics[width=4.0cm]{image3}}
% %  \vspace{1.5cm}
%   \centerline{(b) Results 3}\medskip
% \end{minipage}
% \hfill
% \begin{minipage}[b]{0.48\linewidth}
%   \centering
%   \centerline{\includegraphics[width=4.0cm]{image4}}
% %  \vspace{1.5cm}
%   \centerline{(c) Result 4}\medskip
% \end{minipage}
% %
% \caption{Example of placing a figure with experimental results.}
% \label{fig:res}
% %
% \end{figure}

%\vfill
%\pagebreak

%\vfill\pagebreak
%%% i think the paper looks more full with the bib starting already on the 4th page


\balance

\bibliographystyle{IEEEbib}
\bibliography{refs}

\balance

\end{document}